\documentclass[runningheads]{template/llncs}

\usepackage{amsmath}
\usepackage{amssymb}
\usepackage{bbm}
\usepackage{graphicx}
\usepackage{graphics}
\usepackage[hidelinks]{hyperref}
\usepackage{multirow}

\DeclareMathOperator*{\argmax}{arg\,max}
\DeclareMathOperator*{\argmin}{arg\,min}

\begin{document}

\title{Natural Example-Based Explainability: a Survey}  

\author{
     Antonin Poché$^*$\inst{1,2}\orcidID{0009-0000-0930-3313}
\and Lucas Hervier$^*$\inst{1,2}\orcidID{0000-0002-6642-467X}
\and Mohamed-Chafik Bakkay\inst{1,2}\orcidID{0009-0005-4362-6569}
}

\authorrunning{Poché and Hervier}

\def\thefootnote{*}\footnotetext{These authors contributed equally to this work}
\def\thefootnote{\arabic{footnote}}

\institute{
    IRT Saint Exupéry, Toulouse, France  
    \email{name.surname@irt-saintexupery.com}\\
    \and
    IRT SystemX, 2 boulevard Thomas Gobert, 91120 Palaiseau, France 
    \email{name.surname@irt-systemx.fr}\\
}

\maketitle  

\begin{abstract}

    Explainable Artificial Intelligence (XAI) has become increasingly significant for improving the interpretability and trustworthiness of machine learning models. While saliency maps have stolen the show for the last few years in the XAI field, their ability to reflect models' internal processes has been questioned. Although less in the spotlight, example-based XAI methods have continued to improve. It encompasses methods that use examples as explanations for a machine learning model's predictions. This aligns with the psychological mechanisms of human reasoning and makes example-based explanations natural and intuitive for users to understand. Indeed, humans learn and reason by forming mental representations of concepts based on examples.

    This paper provides an overview of the state-of-the-art in natural example-based XAI, describing the pros and cons of each approach. A "natural" example simply means that it is directly drawn from the training data without involving any generative process. The exclusion of methods that require generating examples is justified by the need for plausibility which is in some regards required to gain a user's trust. Consequently, this paper will explore the following family of methods: similar examples, counterfactual and semi-factual, influential instances, prototypes, and concepts. In particular, it will compare their semantic definition, their cognitive impact, and added values. We hope it will encourage and facilitate future work on natural example-based XAI.

    
    \keywords{Explainability  \and XAI \and Survey \and Example-based \and Case-based \and Counterfactuals \and Semi-factuals \and Influence Functions \and Prototypes \and Concepts}
\end{abstract}

\section{Introduction} \label{introduction}
    
    \begin{figure}
        \centering
        \includegraphics[width=\linewidth]{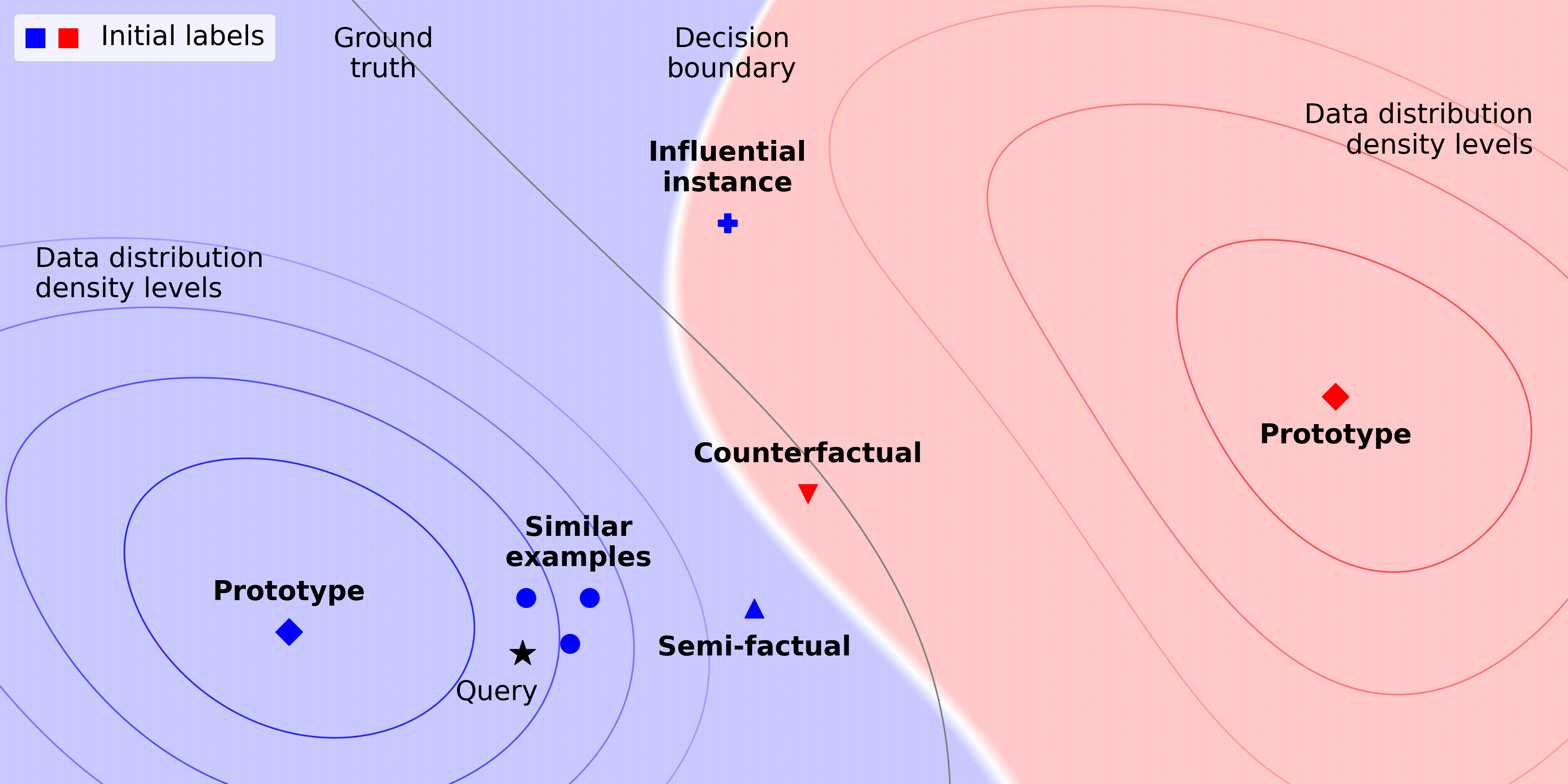}
        \caption{Natural example-based explanation formats w.r.t the studied sample (or query) and the decision boundary. We can see similar examples are the closest elements to the query, while counterfactuals and semi-factual are on either side of the point of the decision boundary the closest to the query. Prototypes are representative of each class in a dense zone of the dataset and the influential instance bends the decision boundary.}
        \label{fig:taxonomy_schema}
    \end{figure}

    With the ever-growing complexity of machine learning models and their large diffusion, understanding models' decisions and behavior became a necessity. Therefore, explainable artificial intelligence (XAI), the field that aims to understand and clarify models, flourished with a huge diversity of methods. To differentiate between methods several taxonomies were proposed, and common components emerged \cite{adadi_2018_peeking,arrieta_2020_explainable,holzinger_2022_explainable}:
    i) Local vs global: Local methods explain a specific decision of the model (in this case, the model's input is called the studied sample or query). Global methods give insight into the general model behavior. Methods can also explain the dataset but it will not be covered in this survey.
    ii) \textit{Post-hoc} vs intrinsic vs explainable by-design: \textit{Post-hoc} methods are applied on an already trained model. By-design methods produce models explainable by construction. Intrinsic methods need to be taken into account during the training of the model but do not affect the final state of the model and explain either the training process or the trained model.
    iii) Black-box vs white-box: White-box methods need access to the model's weights and/or gradients.
    iv) The format of the explanation: The multiplicity of methods translates through the large range of forms of explanations such as attribution methods~\cite{fel_2021_look,selvaraju_2017_grad}, concepts~\cite{fel_2022_craft,kim_2018_interpretability}, surrogate models~\cite{kim2022lightweight,ribeiro_2016_model}, rule-based explanations~\cite{van_der_waa_evaluating_2021}, natural language explanations~\cite{cambria2023survey}, dependencies~\cite{goldstein_2015_peeking,hastie_2009_elements}, and finally example-based explanations~\cite{keane_2019_twin,verma_2022_counterfactual}.

    Nonetheless, no matter the taxonomy of a method, its explanations are aimed at humans, hence, they should exploit the vast literature in philosophy, psychology, and cognitive science on how humans generate, understand, and react to explanations \cite{miller_2019_explanation}. The psychology literature argued that, in everyday life, humans use examples as references to understand, explain something, or demonstrate their arguments \cite{byrne_2016_counterfactual,gentner_1983,miller_2019_explanation,schank_1983}. Afterward, through user studies in the XAI field \cite{fel_2022_craft,humer_comparing_2022,kenny_2021_explaining}, researchers validated that example-based explainability provides better explanations over several other formats. Example-based explainability corresponds to a family of methods where explanations are represented by or communicated through samples, or part of samples like crops. This means the explanation's format is a data point (an example).

    Examples can either be training samples (natural examples) or generated elements. To generate high-dimensional data points, methods are essentially based on deep neural networks \cite{augustin_2022_diffusion,kenny_2021_generating}. However, for most high dimensional data, such methods fail to ensure that generated examples are plausible and belong to the manifold (subspace of the input space where samples follow the data distribution), and examples need to be realistic for humans to interpret them \cite{byrne_2019_counterfactuals}. Therefore, natural examples have two advantages, they do not use a model to explain another model which eases their acceptance, and natural examples are plausible by definition. Hence, this survey will cover natural (non-generative) example-based explainability methods that explain AI models.
    
    Explanations in example-based explainability are all data points but there exist different semantic meanings to a given example. Depending on the relation between the example, the query, and the model, the information provided by the example will differ. The semantic definition of an example and the kind of insight it provides divide the example-based format into sub-groups, which are presented in Fig.~\ref{fig:taxonomy_schema}. This overview is organized around those sub-groups (also called formats), this work will unfold as follows:
    
    The first format is \textbf{similar examples} (or factuals) (Section \ref{similar_examples}), for the model, they are the closest elements to the query. Factuals give confidence in the prediction or explain misclassification, but they are limited to the close range of the considered sample. 
    To provide insight into the model behavior on a larger zone around the query, \textbf{counterfactuals} and \textbf{semi-factuals} (Sections \ref{counterfactuals} and \ref{semifactuals}) are more adapted. They are respectively the closest and the farthest samples on which the model makes a different and similar prediction. They are mainly used in classification, give insight into the decision boundary, and are complementary if paired. While they give an idea of the limit, 
    they do not provide insights on how one could bend the decision boundaries of the model by altering the training data.
    This is addressed through \textbf{influential instances} (Section \ref{influential_instances}), the training samples with the highest impact on the model's state. In addition, contrary to previously listed example-based formats, influential instances are not limited to local explanations. Indeed, one can extract the most influential instances for the model in general. 
    Another global explanation format is \textbf{Prototypes} (Section \ref{prototypes}), which are a set of samples representative of either the dataset or a class. Most of the time they are selected without relying on the model and give an overview of the dataset, but some models are designed through prototypes, thus explainable by design.
    Concepts (Section \ref{concepts}), a closely-related format, is also investigated. A concept is the abstraction of the common elements between samples -- e.g. for trees, the concepts could be trunk, branch, and leaf. To communicate such concepts, if they are not labeled, the easiest way is through examples of such concepts (often part of samples such as patches).
    Finally, \textbf{feature visualization}~\cite{olah2017feature} are generated images that maximize the model prediction for a given class. It shows what the model associate with a given class, however, it is generative and will not be further discussed in this review.

    Thus we could summarize the contributions of this paper as follows:
    i) To the best of our knowledge, we are the first to compile natural example-based explainability literature in a survey. Previous works either covered the whole XAI literature with a superficial analysis of example-based XAI or focused on a given sub-format of example-based XAI.
    ii) For each format we provide simple definitions, semantic meanings, and examples. When possible, we additionally ground formats into social sciences and depict their cognitive added values.
    iii) We explore, classify, and describe available methods in each natural example-based XAI format. We highlight common points and divergences for the reader to understand each method easily, with a focus on key methods. (see Tab.~\ref{tab:methods_comparison})

\subsection{Notations}


Throughout the paper, methods will explain a machine learning model $h: \mathcal{X} \rightarrow \mathcal{Y}$, with $\mathcal{X}$ and $\mathcal{Y}$ being respectively the input and output domain. Especially, this model is parameterized by the weights $\theta \in \Theta \subseteq \mathbb{R}^d$. If not specified otherwise, $h$ is trained on a training dataset $\mathcal{D}_{train} \subset (\mathcal{X} \times \mathcal{Y})$ of size $n$ with the help of a loss function $l: (\mathcal{X}, \mathcal{Y}, \Theta) \rightarrow \mathbb{R}$. We denote a sample by the tuple $z = (x,y) | \quad x \in \mathcal{X}, y \in \mathcal{Y}$. When an index subscript as $i$ or $j$ is added, \textit{e.g.} $z_i$, it is assumed that $z_i$ belongs to the training dataset. If the subscript "test" is added, $z_{test}$, the sample does not belong to the training data. When there is no subscript, the sample can either be or not in the training data. Finally, the empirical risk function is denoted as $\mathcal{L(\theta)} := \frac{1}{n}\sum_{(x, y) \in \mathcal{D}_{train}} l(x, y, \theta) = \frac{1}{n}\sum_{z_{j} \in \mathcal{D}_{train}} l(z_{j}, \theta)$, the parameters that minimized this empirical risk as $\theta^* := \argmin_{\theta} \mathcal{L(\theta)}$ and an estimator of $\theta^*$ is denoted $\hat{\theta}$.

\section{Similar examples} \label{similar_examples}

    In the XAI literature, similar examples, also referred to as factual examples (see Fig.~\ref{fig:contrastive}), are often used as a way to provide intuitive and interpretable explanations. The core idea is to retrieve the most similar, or the closest, elements in the training set to a sample under investigation $z_{test}$ and to use them as a way to explain a model's output. Specifically, Case-Based Reasoning (CBR) is of particular interest as it mimics the way humans draw upon past experiences to navigate novel situations \cite{gentner_1983,schank_1983}. For example, when learning to play a new video game, individuals do not typically begin from a complete novice level. Instead, they rely on their pre-existing knowledge and skills in manipulating game controllers and draw upon past experiences with similar video games to adapt and apply strategies that have been successful in the past. As described by Aamodt and Plaza~\cite{aamodt_1994}, a typical CBR cycle can be delineated by four fundamental procedures: i) RETRIEVE: Searching for the most analogous case or cases, ii) REUSE: Employing the information and expertise extracted from that case to address the problem, iii) REVISE: Modifying the proposed solution as necessary, iv) RETAIN: Preserving the pertinent aspects of this encounter that could be beneficial for future problem-solving endeavors. The CBR approach has gained popularity in fields that require transparent systems to justify their outcomes, such as medicine \cite{bichindaritz_2006}, due to its psychological plausibility. In addition to being intuitive, the cases retrieved by a CBR system for a given prediction are natural explanations for this output.

    While CBR systems are a must-know in the XAI literature, we will not review them as they have already been well analyzed, reviewed, motivated, and described many times \cite{cunningham_2003,demantaras_2005,schoenborn_2021}. Instead, the focus here is on case-based explanations (CBE) \cite{schoenborn_2021}. CBE are methods that use CBR to explain other systems, also referred to as twin systems \cite{keane_2019_twin,kenny_2019_twin}. Indeed, the CBR system must be coupled with the system you want to explain. In particular, explanations of the system under inspection are generally the outcomes of the RETRIEVE functionality of the twinned CBR system, which oftentimes rely on $k$-nearest neighbor ($k$-NN) retrieval \cite{cover_1967}. The idea behind $k$-NN is to retrieve the $k$ most similar training samples (cases) to a test sample $z_{test}$. In fact, presenting similar examples to an end-user as an explanation for a model's outcomes has been shown through user studies to be generally more convincing than other approaches  \cite{jeyakumar_2020,van_der_waa_evaluating_2021}.

    \subsection{Defining similarity}
    Defining similarity is not trivial. Indeed, there are many ways of defining similarity measures, and different approaches are appropriate for different representations of a training sample \cite{demantaras_2005}. Generally, CBR systems assume that similar input features are likely to produce similar outcomes. Thus, using a distance metric defined on those input features engenders a similarity measure: the closer the more similar they are. One of the simplest is the unweighted Euclidean distance:

    \begin{equation}
        \label{eq:cbr:euclidean}
        dist(z, z') = ||x - x'||_2 \quad | \quad z = (x, y) \in (\mathcal{X} \times \mathcal{Y})
    \end{equation}

    However, \textbf{where} -- \textit{i.e.} in which space -- the distance is computed does have major implications. As pointed out by Hanawa \textit{et al.}~\cite{hanawa_evaluation_2021}, the input space does not seem to bring pieces of information on the internal working of the model under inspection but provides more of a data-centric analysis. Thus, recent methods rely instead on either computing the distance in a latent space or weighting features for the $k$-NN algorithm \cite{dudani_1976}.

    \subsubsection{Computing distance in a latent space} is one possibility to include the model in the similarity measure which is of utmost importance if we want to explain it, as pointed out by Caruana \textit{et al.}~\cite{caruana_1999}. Consequently, Caruana \textit{et al.}~\cite{caruana_1999} suggested applying the Euclidean distance on the last hidden units ${h_{-1}}$ of a trained Deep Neural Network (DNN) as a similarity which considers the model's predictions:

    \begin{equation}
        \label{cbr:eq:hidden}
        dist_{DNN}(z, z') = ||h_{-1}(x) - h_{-1}(x')||_2 \quad | \quad z = (x, y) \in (\mathcal{X} \times \mathcal{Y})
    \end{equation}

    Similarly, for Deep Convolutional Neural Networks, Papernot and McDaniel~\cite{papernot_2018}, and Sani \textit{et al.}~\cite{sani_2017} suggested conducting the $k$-NN search in the latent representation of the network and using the cosine similarity distance.

    \subsubsection{Weighting features} is another popular paradigm in CBE. For instance, Shin \textit{et al.}~\cite{shin_2000} proposed various \textbf{global weighting} schemes -- \textit{i.e.} methods in which the weights assigned to each input's feature remain constant across all samples as in Eq.~\eqref{cbr:eq:weight_knn} -- where the weights are computed using the trained network to reveal the input features that were the most relevant for the network's prediction.

    \begin{equation}
        \label{cbr:eq:weight_knn}
        dist_{features\_weights}(z, z') = ||w(\hat{\theta})^T(x-x')||_{2} \quad | \quad z = (x, y) \in (\mathcal{X} \times \mathcal{Y})
    \end{equation}

    Alternatively, Park \textit{et al.}~\cite{park_2004} examined \textbf{local weighting} by considering varying feature weights across the instance space. However, their approach is not \textit{post-hoc} for DNN. Besides, Nugent \textit{et al.}~\cite{nugent_2005} also focused on local weighting and proposed a method that can be applied to any black-box model. However, their method involves generating multiple synthetic datasets around a specific sample, which may not be suitable for explaining a large number of samples or high-dimensional inputs. In the same line of work, Kenny and Keane~\cite{kenny_2019_twin,kenny_2021_explaining} proposed COLE, 
    by suggesting the direct $k$-NN search in the attribution space -- \textit{i.e} computing saliency maps~\cite{bach_2015,shrikumar_2017,sundararajan_2017} for all instances and performing a $k$-NN search in the resulting dataset of attributions. By denoting $c(\hat{\theta}, z)$ the attribution map of the sample $z$ for the model parameterized by $\hat{\theta}$ that gives:

    \begin{equation}
        \label{cbr:eq:cole}
        dist_{COLE}(z, z') = ||c(\hat{\theta}, z) - c(\hat{\theta}, z')||_{2}
    \end{equation}

    They used three saliency map techniques (\cite{bach_2015,shrikumar_2017,sundararajan_2017}) but nothing prevents one to leverage any other saliency map techniques. However, we should also point out that Fel \textit{et al.}~\cite{fel_2022_dont} questioned attribution methods' ability to truly capture the internal process of DNN. Additionally in \cite{kenny_2021_explaining}, Kenny and Keane proposed to use the Hadamard product of the gradient times the input features as a contribution score in the case of DNN with non-linear outputs.

    \subsection{Limitations}
    The current limitations of similarity-based XAI are still significant. Indeed, even though one defines a relevant distance or similarity measure between samples one still has to perform the search in the training dataset to retrieve the closest samples for a given $z_{test}$. Naively, this would at least require computing the distance between $z_{test}$ with every training data point, which prohibits its computation for large datasets. Fortunately, there are efficient techniques available for searching, as briefly discussed in the paper by Bhatia \textit{et al.}~\cite{bhatia_2010}. However, if the training data is sparse in the space in which the distance is computed the retrieved cases might be far from $z_{test}$, thus questioning their relevance.

    Furthermore, \textbf{where} the distance is computed does have major implications as mentioned by Hanawa \textit{et al.}~\cite{hanawa_evaluation_2021}. Consequently, authors have suggested different feature spaces or weighting schemes to investigate, but their relevance to reflect the inner workings of a model is as questionable as it is for attributions methods \cite{fel_2022_dont}. In addition, it is still unclear in the literature if one approach prevails over others. Moreover, when a human is faced with examples, he may not be able to understand why they were considered similar. As an example, if two elements are red and round, the human may think the important thing is the red color while the model focuses on the round shape \cite{nauta2022looks}.

    Finally, the consideration of the position of the retrieved similar examples w.r.t. the decision boundaries of the model, in terms of whether their prediction matches that of $z_{test}$, is not always accounted for. It is a major issue as providing similar examples to an end-user should comfort it with the model's decision but that becomes confusing if you showcase a factual example for which the model's prediction is different. Thus, taking into account the decision boundaries of a model seems crucial for the explanations' relevance. Such considerations are motivating the field of contrastive explanations, as discussed in section \ref{contrastive}.

\section{Contrastive explanations} \label{contrastive}
    
\begin{figure}
        \centering
        \includegraphics[width=\linewidth]{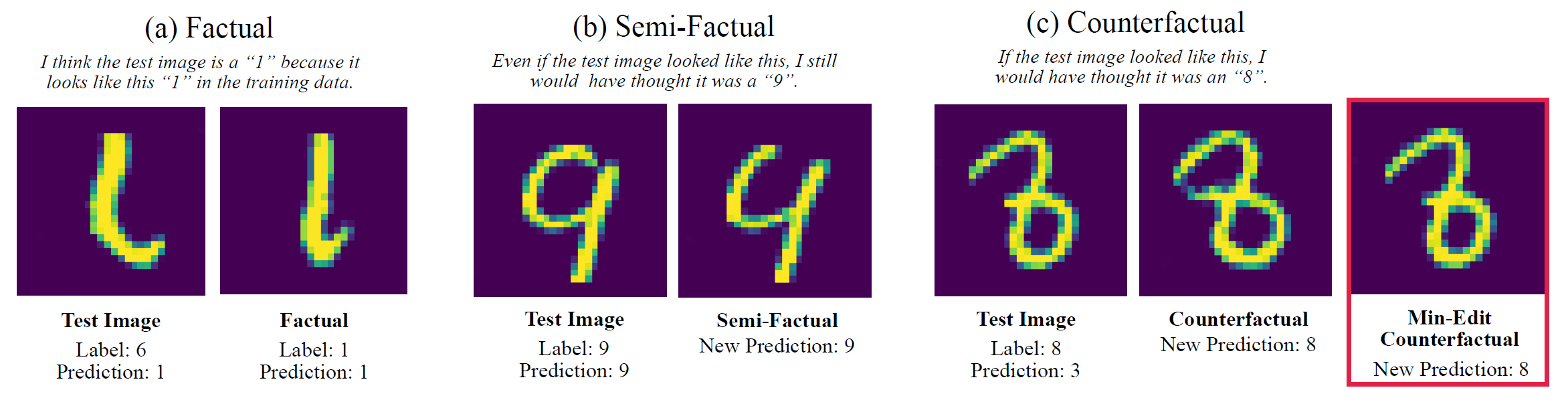}
        \caption{Illustration of factuals, SF, and CF from Kenny and Keane~\cite{kenny_2021_generating}. The factual makes us understand the misclassification, while SF and CF show us how far or close the decision boundary is. Min-edit represents the AP, as differences are not visible.}
        \label{fig:contrastive}
    \end{figure}

    Contrastive explanations are a class of explanation that provides the consequences of another plausible reality, the repercussion of changes in the model's input \cite{byrne_2016_counterfactual,verma_2022_counterfactual}. More simply, they are explanations where we modify the input and observe the reaction of the model's prediction, the modified input is returned as the explanation and its meaning depends on the model's prediction of it. Those methods are mainly \textit{post-hoc} methods applied to classification models. This includes i) counterfactuals (CF): \textit{an imagined alternative to reality about the past, sometimes expressed as “if only ... ” or "what if ..."}~\cite{byrne_2016_counterfactual}, ii) semi-factuals (SF): \textit{an imagined alternative that results in the same outcome as reality, sometimes expressed as “even if ... ”}~\cite{byrne_2016_counterfactual}, and iii) adversarial examples (perturbations or attacks) (AP): \textit{inputs formed by applying small but intentionally worst-case perturbations to examples from the dataset, such that the perturbed input results in the model outputting an incorrect answer with high confidence}~\cite{goodfellow_explaining_2015}. Examples of those three formats are provided in Fig.~\ref{fig:contrastive} from Kenny and Keane~\cite{kenny_2021_generating}.
    
    AP and CF are both perturbations with an expected change in the prediction, they only differ in the goal as CF attempt to provide an explanation of the model's decision while AP are mainly used to evaluate robustness. In fact, AP can be considered CF \cite{wachter_counterfactual_2017}, and for robust models, AP methods can generate interpretable CF \cite{serrurier_2022_adversarial}. Nonetheless, AP are hardly perceptible perturbations designed to fool the model \cite{szegedy_intriguing_2014}, therefore, they are generative and those methods will not be further detailed in this work. Then, we can generalize $SF$ and $CF$, with a given distance $dist$, and the examples conditioned space $\mathcal{X}_{cond(f,x)} \subset \mathcal{X}$:

    \begin{equation}
        \label{eq:cf_definition}
        CF(x_{test}) := \argmin_{x \in \mathcal{X}_{cond(f,x_{test})} | h(x) \neq h(x_{test})}  dist(x_{test},x)
    \end{equation}
    
    \begin{equation}
        \label{eq:sf_definition}
        SF(x_{test}) := \argmax_{x \in \mathcal{X}_{cond(f,x_{test})} | h(x) = h(x_{test})}  dist(x_{test},x)
    \end{equation}

    For natural CF and SF, the input space is conditioned to the training set, $\mathcal{X}_{cond(f,x_{test})} = X_{train}$. While for AP, there is no condition on the input space, in Eq.~\eqref{eq:cf_definition}, $\mathcal{X}_{cond(f,x_{test})} = \mathcal{X}$. The distance and the condition of the input space are the key differences between CF and SF methods.

    \subsection{Counterfactuals} \label{counterfactuals}    
            
        \subsubsection{The social science grounding} \label{cf_grounding}
            of counterfactuals is deep, either in philosophy, or psychology. Indeed, the search for CF's semantic definition goes back a long time \cite{bennett_2003_philosophical,halpern_2005_explanations,lewis_1973_counterfactuals}, and historically revolves around the notion of cause and effect, sometimes called facts and foils \cite{lipton_1990_contrastive,miller_2019_explanation}. Then, Halpern and Pearl~\cite{halpern_2005_explanations} proved that the cause of an event is an answer to the question "Why?" and thus, provides a powerful explanation. Moreover, the philosophical literature argued that CF allow us to communicate and understand the causal relation between facts and foils \cite{lewis_1973_counterfactuals,miller_2019_explanation}.
            Psychology also possesses a rich literature regarding CF \cite{byrne_2016_counterfactual,roese_1995_counterfactual}, which has continued to evolve in recent years \cite{byrne_2019_counterfactuals,keane_2020_good,miller_2021_contrastive} thanks to the arrival of CF in XAI through Wachter \textit{et al.}~\cite{wachter_counterfactual_2017}. Humans' natural use of counterfactuals in many situations was highlighted by Byrne~\cite{byrne_2016_counterfactual}: \textit{From amusing fantasy to logical support, they explain the past, prepare the future, modulate emotional experience, and support moral judgments}. Furthermore, when people encounter CF they have both the counterfactual and the factual in mind \cite{byrne_2019_counterfactuals,thompson_2002_reasoning}.
            The insights from philosophy and psychology \cite{byrne_2019_counterfactuals,miller_2021_contrastive} have shown the pertinence and potential of CF as well as SF for XAI. To match such promises, CF in XAI need to verify the definitions and properties of CF typically employed by humans.

        \subsubsection{Expected properties} \label{cf_properties}
            for natural CF can be extrapolated from conclusions and discovered properties in XAI for generated CF even though the literature on natural CF is slim. Such desirable properties for CF, derived from social sciences, could be summarized as follows:
            i) \textbf{plausibility}~\cite{keane_2021_if,keane_2020_good,verma_2022_counterfactual}: CF should be as realistic as possible;
            ii) \textbf{validity}~\cite{mothilal_2020_explaining}: if the model's prediction on CF differ from the prediction on the query (see the definition \eqref{eq:cf_definition}); 
            iii) \textbf{sparsity}~\cite{keane_2021_if,mothilal_2020_explaining,verma_2022_counterfactual}: the number of features that were changed between CF and the query should be as little as possible;
            iv) \textbf{diversity}~\cite{karimi_2020_modelagnostic,mothilal_2020_explaining}: if several CF are proposed, they should be different from each other;
            v) \textbf{actionability}~\cite{keane_2021_if,verma_2022_counterfactual}: the method should allow the user to select features, to modify and specify immutable ones;
            vi) \textbf{proximity}~\cite{karimi_2020_modelagnostic,keane_2021_if,keane_2020_good,mothilal_2020_explaining}: CF should be as close as possible to the query.

        \subsubsection{Counterfactuals methods:} \label{cf_methods}
            Keane \textit{et al.}~\cite{keane_2020_good} argued that nearest unlike neighbors (NUN)~\cite{dasarathy_1995_nearest} is the ancestor of counterfactuals in XAI. NUN are derivative of nearest neighbors \cite{cover_1967}, which is looking for the nearest element that belongs to a different class, it matches perfectly with the definition of natural counterfactuals. NUN were first used in XAI by Doyle \textit{et al.}~\cite{doyle_2004_explanation,nugent_2009_gaining} but not as an explanation, only to find SF. The only method to the best of our knowledge that uses NUN as explanations is KLEOR from Cummins and Bridge~\cite{cummins_2006_kleor}, which was also called "the nearest miss" and was provided as a complement to SF explanation. Indeed, following the definition, pairs of CF and SF should give a good intuition of the decision boundary. Nonetheless, they highlighted that the decision boundary might be much more complex than what the SF and CF pairs can reveal. Indeed, a line between SF and CF may intersect the decision boundary several times, which can lead to explanations that are not always faithful. Furthermore, Keane \textit{et al.}~\cite{keane_2020_good} argued that "good natural counterfactuals are hard to find" as the dataset's low density prevents sparse and proximal natural CF.

            Counterfactuals as known in XAI appeared with Wachter \textit{et al.} \cite{wachter_counterfactual_2017}. While there are numerous methods, as shown through the number of surveys in this field\cite{karimi_2020_modelagnostic,mothilal_2020_explaining,verma_2022_counterfactual}, those are all generative methods. We can distinguish two periods among those papers: a first one with a focus on small and interpretable tabular datasets as described by Verma \textit{et al.} survey~\cite{verma_2022_counterfactual}, and a second on more complex data types such as images \cite{augustin_2022_diffusion,kenny_2021_generating}. While in the first CF period, generating plausible instances was not an issue, it appeared to be a huge drawback toward the generalization of CF to more complex data types \cite{augustin_2022_diffusion,kenny_2021_generating}. Even the most recent methods based on diffusion models \cite{augustin_2022_diffusion} failed to consistently generate plausible images. We are surprised that there is so little work that explores natural CF as explanations with their inherent plausibility. Furthermore, in the literature, natural examples were used to ensure plausibility in generated CF \cite{keane_2020_good,verma_2022_counterfactual}.
            Moreover, adversarial perturbations proved that for non-robust DNN, a generated example close to a natural instance is not necessarily plausible. That is to say, we cannot prove that generated instances belong to the manifold without a proper definition of the manifold. To conclude, for high dimensional data, the reader is faced with the choice of simple and plausible natural CF or proximal and sparse generated CF through a model explaining another model.

    \subsection{Semi-factuals} \label{semifactuals}
    
        SF literature is most of the time included in the CF literature be it in philosophy \cite{goodman_1947_problem}, psychology \cite{byrne_2016_counterfactual}, or XAI \cite{cummins_2006_kleor,kenny_2021_generating}. In fact, SF, "even if ..." are semantically close to CF, "what if ..."~\cite{aryal_2023_even,bennett_2003_philosophical,goodman_1947_problem}, (see Eqs.~\eqref{eq:cf_definition} and \eqref{eq:sf_definition}). However, psychology has demonstrated that human reactions differ between CF and SF. While CF strengthen the causal link between two elements, SF reduce it \cite{byrne_2019_counterfactuals}, CF increase fault and blame in a moral judgment while SF diminish it.

        \subsubsection{Expected properties} \label{sf_properties}
            for CF and SF were inspired by social science, hence, because of their close semantic definition, many properties are common between both: SF should also respect their definition in Eq.~\eqref{eq:sf_definition} (\textbf{validity}), then to make the comparison possible and relevant they should aim towards \textbf{plausibility}~\cite{aryal_2023_even}, \textbf{sparsity}~\cite{aryal_2023_even}, \textbf{diversity}, and \textbf{actionability}. Nonetheless, the psychological impact of CF and SF differ, hence there are also SF properties that contrast with CF properties. The difference between equations \eqref{eq:cf_definition} and \eqref{eq:sf_definition} -- \textit{i.e.} $\argmin$ vs $\argmax$ -- suggests that to replace CF's proximity, SF should be the farthest from the studied sample, while not crossing the decision boundary  \cite{cummins_2006_kleor}. As such, we propose the \textbf{decision boundary closeness} as a necessary property, and a metric to evaluate it could be the distance between SF and SF's NUN. Finally, SF should not go in any direction from the studied sample but aim toward the closest decision boundary. Therefore, it should be aligned with NUN \cite{cummins_2006_kleor,doyle_2004_explanation,nugent_2009_gaining}, this property was not named, we suggest calling it \textbf{counterfactual alignment}.

        \subsubsection{Semi-factuals methods} \label{sf_methods}
            were first reviewed in XAI by a recent survey from Aryal and Keane~\cite{aryal_2023_even}. They divided SF methods and history into four parts. The first three categories consist of one known method that will illustrate them:
            \begin{itemize}
                \item \textbf{SF based on feature-utility}, Doyle \textit{et al.} \cite{doyle_2004_explanation} discovered that similar examples may not be the best explanations and suggested giving examples farther from the studied sample. To find the best explanation case, $dist$ in Eq.~\eqref{eq:sf_definition} is a utility evaluation based on features difference.
                \item \textbf{NUN-related SF}, Cummins and Bridge~\cite{cummins_2006_kleor} proposed KLEOR where Eq.~\eqref{eq:sf_definition}'s $dist$ is based on NUN similarity. Then, they penalize this distance to make sure the SF are between the query and nearest unlike neighbors.
                \item \textbf{SF near local-region boundaries}, Nugent \textit{et al.}~\cite{nugent_2009_gaining} approximate the decision boundary of the model in the neighborhood of the studied sample through input perturbations (like LIME~\cite{ribeiro_2016_model}). Then SF are given by the points that are the closest to the decision boundary.
                \item \textbf{The modern era: \textit{post}-2020 methods}, inspired by CF methods, many generative methods emerged in recent years \cite{karras_2020_analyzing,kenny_2021_generating}.
            \end{itemize}

            In conclusion, semi-factuals are a natural evolution of similar examples. Furthermore, their complementarity with counterfactuals was exposed through the literature, first to find and evaluate SF, and then to provide a range to the decision boundary.
            Even though contrastive explanations bring insights into a model's behavior by answering a "\textit{what if...} or a "\textit{even if...}" statement, it has no impact on the current model situation and what led to this state or how to change it. Contrastively, influential instances (see Section \ref{influential_instances}) extract the samples with the most influence on the model's training, hence its current state. Thus, removing such samples from the training set will have a huge impact on the resulting model.

\section{Influential Examples} \label{influential_instances}

    Influential instances could be defined as instances more likely to change a model's outcome if they were not in the training dataset. Furthermore, such measures of influence provide one with information on "in which direction" the model decision would have been affected if that point was removed. Being able to trace back to the most influential training samples for a given test sample $z_{test}$ has been a topic of interest mainly for example-based XAI.

    \subsection{Influence functions}
        \subsubsection{Influence functions} originated from robust statistics in the early 70s. In essence, they evaluate the change of a model's parameters as we up-weight a training sample by an infinitesimal amount: \cite{hampel_1974} $\hat{\theta}_{\epsilon, z_j} := \argmin_{\theta} \mathcal{L}(\theta) + \epsilon l(z_j, \theta)$. One way to estimate the change in a model's parameters of a single training sample would be to perform \textit{Leave-One-Out} (LOO) retraining, that is, to train the model again with the sample of interest being held out of the training dataset. However, repeatedly re-training the model to exactly retrieve the parameters' changes could be computationally prohibitive, especially when the dataset size and/or the number of parameters grows. As removing a sample $z_j$ can be linearly approximated by up-weighting it by $\epsilon = -\frac{1}{n}$, computing influence helps to estimate the change of a model's parameters if a specific training point was removed. Thus, by making the assumption that the empirical risk $\mathcal{L}$ is twice-differentiable and strictly convex w.r.t. the model's parameters $\theta$ making the Hessian $H_{\hat{\theta}} := \frac{1}{n}\sum_{z_{i} \in \mathcal{D}_{train}}\nabla^{2}_{\theta}l(z_i, \hat{\theta})$ positive definite, Cook and Weisberg~\cite{cook_1982} proposed to compute the influence of $z_j$ on the parameters $\hat{\theta}$ as:
    
        \begin{equation}
            \label{if:self_influence}
            \mathcal{I}(z_j) := -H_{\hat{\theta}}^{-1}\nabla_{\theta}l(z_j,\hat{\theta})
        \end{equation}
    
        Later, Koh and Liang~\cite{koh_2017} popularized influence functions in the machine learning community as they took advantage of auto-differentiation frameworks to efficiently compute the hessian for DNN and derived Eq.~\eqref{if:self_influence} to formulate the influence of up-weighting a training sample $z_j$ on the loss at a test point $z_{test}$:
    
        \begin{equation}
            \label{if:eq:influence_function}
            \mathrm{IF}(z_j, z_{test}) := -\nabla_{\theta}l(z_{test}, \hat{\theta})^{T} H_{\hat{\theta}}^{-1}\nabla_{\theta}l(z_j,\hat{\theta})
        \end{equation}
    
        This formulation opens its way into example-based XAI as it compares to the study of finding the nearest neighbors of $z_{test}$ in the training dataset -- \textit{i.e.} the most similar examples (Section \ref{similar_examples}) -- with two major differences though: i) points with high training loss are given more influence \textit{revealing that outliers can dominate the model parameters}~\cite{koh_2017}, and ii) $H_{\hat{\theta}}^{-1}$ measures what Koh and Liang called: \textit{the resistance of the other training points to the removal of $z_j$}~\cite{koh_2017}. However, it should be noted that hessian computation remains a significant challenge, that could be alleviated with common techniques \cite{agarwal_2017,martens_2010,schioppa_2021}. By normalizing Eq.~\eqref{if:eq:influence_function}, Barshan \textit{et al.}~\cite{barshan_2020} further added stability to the formulation.
    
        Oftentimes, we are not only interested in individual instance influence but in the influence of a group of training samples (\textit{e.g.} mini-batch effect, multi-source data, etc..). Koh \textit{et al.}~\cite{koh_2019} suggested that using the sum of individual influences as the influence of the group constitutes a good proxy to rank those groups in terms of influence. Basu \textit{et al.} \cite{basu_2019} on their side suggested using a second-order approximation to capture possible cross-correlations but they specified it is most likely impracticable for DNN. In a later work, Basu \textit{et al.} \cite{basu_2020} concluded that influence function estimates for DNN are fragile as the assumptions on which they rely, being near optimality and convexity, do not hold in general for DNN.
    
        \subsubsection{LOO approximation} is one of the previously mentioned motivations behind influence estimates as it avoids the prohibitive LOO retraining required for every sample in the training data. Thus, some authors proposed approaches that optimize the number of LOO retraining necessary to get a grasp on a sample's influence such as Feldman and Zhang~\cite{feldman_2020}. Although this significantly reduces the number of retraining compared to naive LOO retraining, it still requires a significant amount of them. Recently, a new approach that relates to influence functions and involves training many models, was introduced with data models \cite{ilyas_2022,saunshi_2022} which we do not review here. 
        
        As Basu \textit{et al.}~\cite{basu_2020} pointed out, there is a discrepancy between LOO approximation and influence function estimates, especially for DNN. However, Bae \textit{et al.}~\cite{bae_2022} claimed that this discrepancy is due to influence functions approaching what they call the proximal Bregman response function (PBRF), rather than approximating the LOO retraining, which does not interfere with their ability to perform the task they were thought for, especially XAI. Thus, they suggested evaluating the quality of influence estimates by comparing them to the PBRF rather than LOO retraining as it was done until now.   

    \subsection{Other techniques}
        \subsubsection{Influence computation that relies on kernels} is another paradigm to find the training examples that are the most responsible for a given set of predictions. For instance, Khanna \textit{et al.}~\cite{khanna_2018} proposed an approach that relies on Fisher's kernels and they related it to the one from Koh and Liang~\cite{koh_2017} as a generalization of the latter under certain assumptions. Yeh \textit{et al.}~\cite{yeh_2018} also suggested an approach that leverages kernels but this time they relied on the representer theorem~\cite{scholkopf_2001}. That allows them to focus on explaining only the \textit{pre-activation prediction layer} of a DNN for classification tasks. In addition, their influence scores, called representer values, provide supplementary information, with positive representer values being excitatory and negative values being inhibitory. However, this approach requires introducing an $L2$ regularizer during optimization, which can prevent \textit{post-hoc} analysis if not responsible for training. Additionally, Sui \textit{et al.}~\cite{sui_2021} argued that this approach provides more of a \textit{class-level} explanation rather than an \textit{instance-level} explanation. To address this issue and the $L2$ regularizer problem, they proposed a method that involves hessian computation on the classification layer, with only the associated computational cost. However, the ability to retrieve relevant samples when investigating only the final prediction layer was questioned by Feldmann and Zhang~\cite{feldman_2020}, who found that memorization does not occur in the last layer.
    
        \subsubsection{Tracing the training process} has been another research field to compute influence scores. It relies on the possibility to replay the training process by saving some checkpoints of our model parameters, or states, and reloading them in a post-hoc fashion \cite{chen_2021,hara_2019,pruthi_2020}. In contrast to the previous approaches, they rely neither on being near optimality nor being strongly convex, which is more realistic when we consider the reality of DNN. However, they require handling the training procedure to save the different checkpoints, potentially numerous, hence they are intrinsic methods, which in practice is not always feasible.

    \subsection{In a nutshell}
\begin{figure}[ht]
        \centering
        \includegraphics{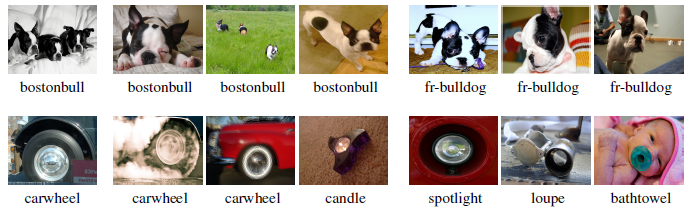}
        \caption{Figure taken from F. Liu \cite{pruthi_2020}: A tracing process for estimating influence, $\textrm{TracIn}$, applied on ImageNet. The first column is composed of the test sample, the next three columns display the training examples that have the most positive value of influence score while the last three columns point out the training examples with the most negative values of influence score. (fr-bulldog: french-bulldog)}
        \label{if:fig:tracin}
    \end{figure}

    Influential techniques can provide both global and local explanations to enhance model performance. Global explanations allow for the identification of training samples that significantly shape decision boundaries or outliers (see Fig.~\ref{fig:taxonomy_schema}), aiding in data curation. On the other hand, local explanations offer guidance for altering the model in a desired way (see Fig.~\ref{if:fig:tracin}). Although they have been compared to similar examples and have been shown to be more relevant to the model \cite{hanawa_evaluation_2021}, they are more challenging to interpret and their effectiveness for trustworthiness is unclear. Further research, particularly user studies, is necessary to determine their ability to take advantage of human cognitive processes.

\section{Prototypes} \label{prototypes}

  \begin{figure}[ht]
        \centering
        \includegraphics[width=0.95\linewidth]{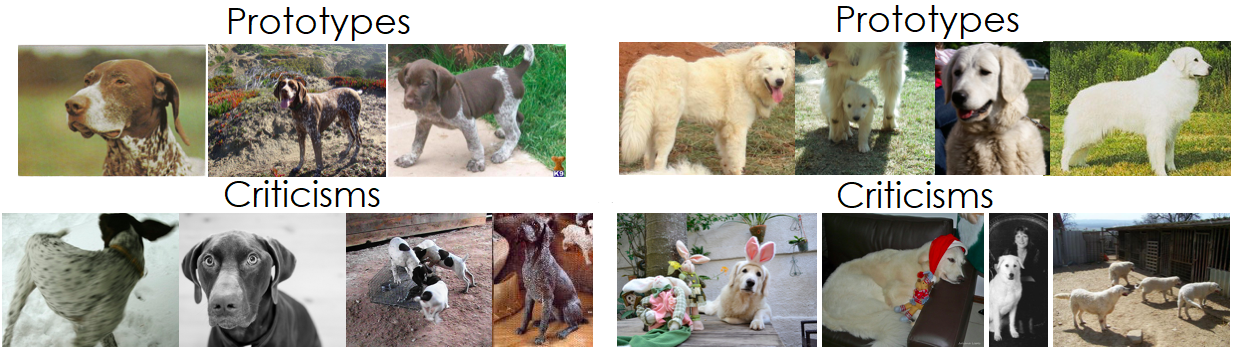}
        \caption{Figure taken from \cite{kim_examples_2016}: Learned prototypes and criticisms from Imagenet dataset (two types of dog breeds)}
        \label{fig:prototypes}
    \end{figure}
    
    Prototypes are a set of representative data instances from the dataset, while criticisms are data instances that are not well represented by those prototypes~\cite{kim_examples_2016}. Fig.~\ref{fig:prototypes} shows examples of prototypes and criticisms from Imagenet dataset. Prototypes and criticism can be used to add data-centric interpretability, \textit{post-hoc} interpretability, and to build an interpretable model~\cite{molnar2020interpretable}. The data-centric approaches will be very briefly introduced.

    \subsection{Prototypes for data-centric interpretability}
        Clustering algorithms that return actual data points as cluster centers such as k-medoids methods \cite{kaufman1990finding,ng1994cient} could be used to better understand the data distribution. In fact, the cluster centers can be considered as prototypes. 
    
        The abundance of large datasets has renewed the interest in the data summarization methods~\cite{badanidiyuru2014streaming,lin2010multi,lin2011class,mirzasoleiman2015distributed,simon2007scene}, also known as set cover methods, which consist of finding a small subset of data points that covers a large dataset. The subset elements can be considered prototypes. Additionally, we found data summarization methods based on the Maximum Mean Discrepancy (MMD), such as MMD-critic~\cite{kim_examples_2016} and Protodash \cite{gurumoorthy_efficient_2019}, that learn both prototypes and criticisms.

    \subsection{Prototypes for \textit{post-hoc} interpretability}
        Prototypes and criticisms can be used to add \textit{post-hoc} interpretability~\cite{molnar2020interpretable}. This can be achieved by predicting the outputs for the selected prototypes and criticisms with the black-box model, and then using these predictions to find the weaknesses of the model. We can also explain the model by applying clustering and data summarization methods to select prototypes in its latent space.
        
        Filho \textit{et al.}~\cite{filho2023explainable} proposed M-PEER (Multiobjective Prototype-based Explanation for Regression) method that finds the prototypes using both the training data and the model output. It optimizes both the error of the explainable model and the fidelity and interpretability metrics. The selected prototypes are then used to provide global and local post-hoc explanations for regression problems.

    \subsection{Prototype-based models interpretable by design}
        After data-centric and \textit{post-hoc} methods, there are methods that construct prototype-based models. Those models are interpretable by design because they provide a set of prototypes that make sense for the model, those methods are mainly designed for classification. Given a training set of points $X_c := \{(x,y) \in \mathcal{D}_{train} | y = c\}$ for each class $c$, an interpretable classifier learns a set of prototypes $P_c \subseteq X_c$ for each class $c$. Each $P_c$ is designed to capture the full variability of the class $c$ while avoiding confusion with other classes. The learned prototypes are then used by the model to classify the input. We identified three types of prototype-based classifiers: those that resolve set cover problems, those that use Bayesian models for explanation, and those that are based on neural networks.

        \subsubsection{Prototype-based classifiers resolving set cover problems} 
            select convex sets that cover each class with prototypes to represent it. Various types of convex sets such as boxes, balls, convex hulls, and ellipsoids can be used. Class Cover Catch Digraphs (CCCD)~\cite{Priebe2003classification} and ProtoSelect~\cite{bien2011prototype} used balls where the centers were considered prototypes. Then, the nearest-prototype rule is used to classify the data points. CCCD finds, for each class $c$, one ball that covers all points of the class $c$ and no points of other classes. Its radius is chosen as large as possible. However, even within large classes, there can still be a lot of interesting within-class variability that should be taken into account when selecting the prototypes. To overcome this limitation, ProtoSelect used a fixed radius across all points, to allow the selection of multiple prototypes for large classes, and they also allow wrongly covered and non-covered points. They simultaneously minimize three elements: i) the number of prototypes; ii) the number of uncovered points; iii) the number of wrongly covered points.

        \subsubsection{Prototype-based classifiers using Bayesian models for explanation:}
            Kim \textit{et al.} ~\cite{kim_bayesian_2015} proposed the Bayesian Case Model (BCM) that extends Latent Dirichlet Allocation ~\cite{blei2003latent}. In BCM, the idea is to divide the data into $s$ clusters. For each cluster, a prototype is defined as the sample that maximizes the subspace indicators that characterize the cluster. When a sample is given to BCM, this last one yield a vector of probability to belong to each of the $s$ clusters which can be used for classification. Thus, the classifier uses as an input a vector of dimension $s$, which allows the use of simpler models due to dimensionality reduction. In addition, the prototype of the most likely cluster can then be used as an explanation.

        \subsubsection{Prototype-based neural network classifiers}
            learn to select prototypes defined in the latent space, which are used for the classification. This lead to a model that is more interpretable than a standard neural network since the reasoning process behind each prediction is ``transparent''. Learning Vector Quantization (LVQ)~\cite{kohonen1990self} is widely used for generating prototypes as weights in a neural network. However, the use of generated prototypes reduces their interpretability. ProtoPNet~\cite{chen2019looks} also stocks prototypes as weights and trains them, but projects them to training samples patches representation during training. Given an input image, its patches are compared to each prototype, the resulting similarity scores are then multiplied by the learned class connections of each prototype. ProtoPNet has been extended to time series data using ProSeNet~\cite{ming2019interpretable}, or with a more interpretable structure with ProtoTree~\cite{nauta2021neural} and HPNet~\cite{hase2019interpretable}. Instead of using linear bag-of-prototypes, ProtoTree and HPNet used hierarchically organized prototypes to classify images. ProtoTree improves upon ProtoPNet by using a decision tree which provides an easy-to-interpret global explanation and can be used to locally explain a single prediction. Each node in this tree contains a prototype (as defined by ProtoPNet). The similarity scores between image patches and the prototypes are used to determine the routing through the tree. Decision-making is therefore similar to human reasoning \cite{nauta2021neural}. Nauta \textit{et al.}~\cite{nauta2022looks} proposed a method called ``This Looks Like That, Because'' to complete the ``This Looks Like That'' reasoning used in ProtoPnet. This method allows checking why the model considered two examples as similar. For instance, it is possible that a human thinks that the common point between two examples is their color, while the model uses their shape. The method modifies some characteristics of the input image, such as hue, or shape, to observe how the similarity score changes. This allows us to measure the importance of each of these characteristics.

    \subsection{In conclusion}
    
        Prototypes can either be: 
        i) selected from the training data to explain the data distribution. These prototypes can also be used to find weaknesses of a black-box model by analyzing the output prediction of these prototypes with this model. 
        ii) selected using both the training data and the model output or in the latent space of the model. This allows for \textit{post-hoc} explanations on the model. 
        iii) integrated and selected by the model itself during training and then used for prediction. This allows the model to be interpretable by design.

\section{Concept-based XAI} \label{concepts}
\begin{figure}
        \centering
        \includegraphics[width=0.95\linewidth]{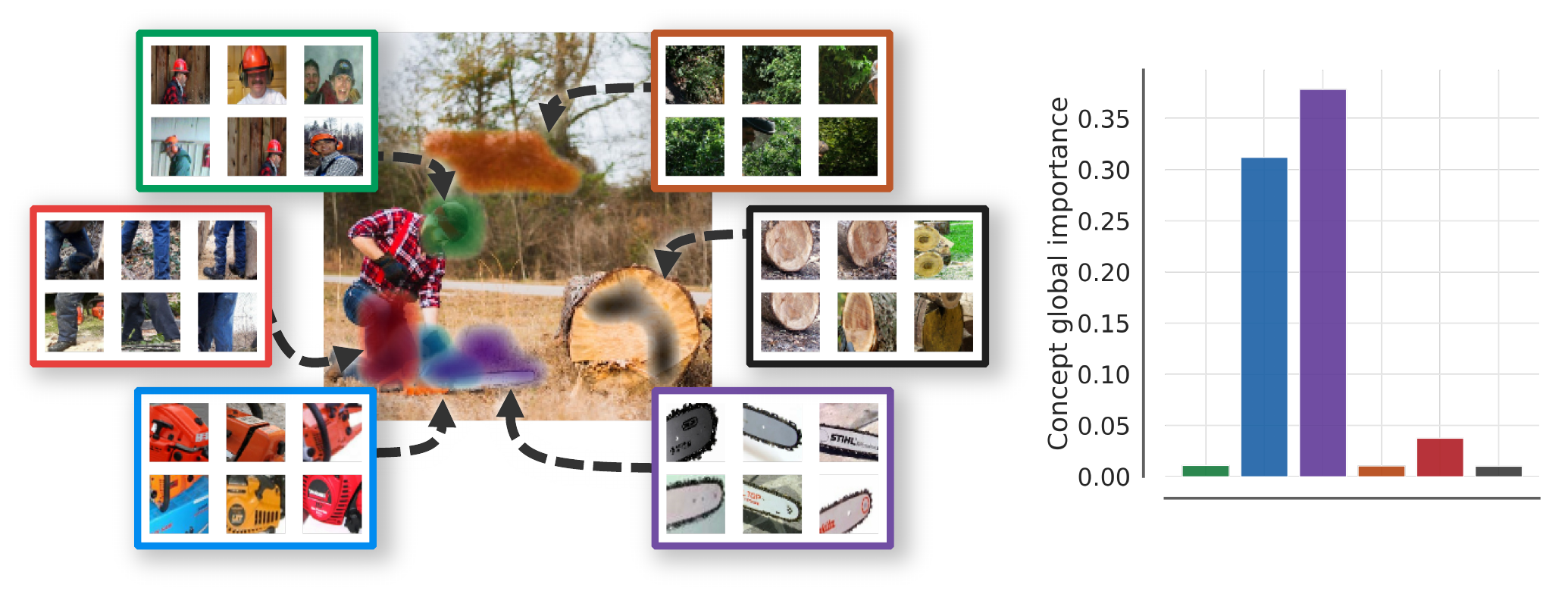}
        \caption{Illustration from Fel \textit{et al.}~ \cite{fel_2022_craft}. Natural examples in the colored boxes define a concept. \textbf{Purple box}: could define the concept of "\textbf{chainsaw}". \textbf{Blue box}: could define the concept of "\textbf{saw's motor}". \textbf{Red box}: could define the concept of "\textbf{jeans}".}
        \label{fig:cbr:craft}
    \end{figure}

    Prototype-based models compare prototypical parts, \textit{e.g.} patches, and the studied sample to make the classification. The idea of parts is not new to the literature, the part-based explanation field, developed for fine-grained classification, is able to detect semantically significant parts of images. The first part-based model required labeled parts for training and can be considered object detection with a semantic link between the detected objects. Afterward, unsupervised methods such as OPAM~\cite{peng2017object} or Particul~\cite{xu2022particul} emerged, those methods still learned classification in a supervised fashion, but no labels were necessary for part identification. In fact, the explanation provided by this kind of method can be assimilated into concept-based explanations. A concept is an abstraction of common elements between samples, as an example Fig. \ref{fig:cbr:craft} shows the visualization of six different concepts that the CRAFT method \cite{fel_2022_craft} associated with the given image. To understand parts or concepts, the method uses examples and supposes that with a few examples, humans are able to identify the concept.

    Like in part-based XAI, the first concept-based method used labeled concepts. Kim et al. \cite{kim_2018_interpretability} introduced concept activation vectors (CAV) to represent concepts using a model latent space representation of images. Then, they design a post-hoc method, TCAV \cite{kim_2018_interpretability} based on CAV to evaluate an image correspondence to a given concept. Even though it seems promising, this method requires prior knowledge of the relevant concepts, along with a labeled dataset of the associated concepts, which is costly and prone to human biases.

    Fortunately, recent works have been conducted to automate the concept discovery in the training dataset without humans in the loop. For instance, ACE, proposed by Ghobarni et al. \cite{ghorbani_2019_towards}, employs a semantic segmentation technique on images belonging to a specific class of interest and use an Inception-V3 neural network to compute activations of an intermediate model layer for these segments. The resulting activations are then clustered to form a set of prototypes, which they refer to as "concepts". However, the presence of background segments in these concepts requires a post-processing clean-up step to remove irrelevant and outlier concepts. Zhang et al. \cite{zhang_2021_invertible} propose an alternative approach to solve the unsupervised concept discovery problem through matrix factorizations \cite{lee_1999_learning} in the networks' latent spaces. However, such methods operate at the convolutional kernel level, which may lead to concepts based on shape and/or ignore more abstract concepts.
    
    As an answer, Fel et al. \cite{fel_2022_craft} propose CRAFT, which uses Non-Negative Matrix Factorization \cite{lee_1999_learning} for concept discovery. In addition to filling in the blank of previous approaches, their method provides an explicit link between the concepts' global and local explanations (Fig. \ref{fig:cbr:craft}). While their approach successfully alleviates the previously mentioned issues, the retrieved concepts are unfortunately not always interpretable. Nonetheless, their user study proved the pertinence of the method.

    To conclude, concept-based explanations allow \textit{post-hoc} global and local explanations, by understanding the general concepts associated with a given class and the concepts used for a decision. We draw attention to methods that do not require expert knowledge to find out relevant concepts as it is prone to confirmation bias. Even though automated concept discovery is making tremendous progress, the interpretation of such concepts and their ability to gain users' trust stay questionable as very few user studies have been conducted on the subject.

\section{Conclusions}

    This paper explored explainability literature about natural example-based explainability and provided a general social science justification for example-based XAI. We described each kind of explanation possible through samples. For each possibility, we reviewed what explanation do they bring, classified and presented the major methods. We summarize all explored described methods in table \ref{tab:methods_comparison}. We saw that all those methods are based on a notion of similarity. As such, for them to explain the model, the similarity between instances should take into account the model. There are two ways of doing it: project the instances in a meaningful space for the model and/or weight instances.

    Among the formats, similar examples and influential instances are natural examples by definition. However, contrastive explanations, prototypes, and concept examples can be generated, which brings competition to non-generative methods. We argue that while a "good" natural example may not exist for a given case, at least, natural examples are realistic in the sense that they belong to the data distribution. While generative methods may be able to create such "good" examples, they cannot prove that the generated samples belong to the data manifold. Furthermore, such methods require a model to explain another model, which in turn should be investigated and might involve extensive tuning.

    We have illustrated that the different example-based formats bring different kinds of explanations, and each one has its own advantages, Fig.~\ref{fig:taxonomy_schema} shows their diversity and complementarity. To summarize those advantages non-exhaustively: i) Factuals give confidence in the decisions of the model and are pertinent in AI-assisted decisions. ii) For classification, contrastive explanations give insight into the decision boundary in the locality of the studied sample. iii) Influential instances explain how samples influenced the model training. iv) Prototypes and concepts give information on a global scale, on the whole, model behavior, but may also be used to explain decisions. Nonetheless, like all explanations, we cannot be sure that humans will have a correct understanding of the model or the decision. Furthermore, there is a non-consensus on how to ensure a given method indeed explain the decisions or inner working of the model. Moreover, for example-based explainability, the data is used as an explanation, hence, without profound knowledge of the dataset, humans will not be able to draw conclusions through such explanations. Therefore, the evaluation of example-based methods should always include a user study, which is lacking in this field and in XAI in general. Finally, we hope our work will motivate, facilitate and help researchers to keep on developing the field of XAI and in particular, natural example-based XAI and to address the identified challenges.

    \begin{table}[tbh]
    \footnotesize
    \setlength\doublerulesep{0.3cm} 
    \centering
    \resizebox{\columnwidth}{!}{
    \begin{tabular}{|c|c|c|c|c|c|c|}
        \hline
        \textbf{SIMILAR} & \multirow{2}{*}{Year} & Global / & Post-hoc / & Model or data & \multirow{2}{*}{Distance} & \multirow{2}{*}{Weighting}\\
        \textbf{EXAMPLES} & & Local & Intrinsic & -type specificity & & \\
        \hline
        Caruana et al. \cite{caruana_1999} & 1999 & Local & Post-hoc & DNN & Euclidean & None \\ \hline
        Shin et al. \cite{shin_2000} & 2000 & Local & Post-hoc & DNN & Euclidean & Global \\ \hline
        Park et al. \cite{park_2004} & 2004 & Local & Intrinsic & DNN & Euclidean & Local \\ \hline
        Nugent et al. \cite{nugent_2005}& 2005 & Local & Post-hoc & None & Euclidean & Local \\ \hline
        Sani et al. \cite{sani_2017} & 2017 & Local & Post-hoc & Deep CNN & Cosine similarity & Local \\ \hline
        Papernot and McDaniel \cite{papernot_2018} & 2018 & Local & Post-hoc & Deep CNN & Cosine similarity & Local \\ \hline
        Cole \cite{kenny_2019_twin} \cite{kenny_2021_explaining} & 2019 & Local & Post-hoc & None & Euclidean & Local with attributions \\ \hline
        \hline
        \textbf{CONTRASTIVE} & \multirow{2}{*}{Year} & Global / & Post-hoc / & Model or data & \multicolumn{2}{c|}{Semi-factual}\\
        \textbf{EXPLANATIONS} & & Local & Intrinsic & -type specificity & \multicolumn{2}{c|}{group of method}\\
        \hline
        Doyle et al. \cite{doyle_2004_explanation,doyle_2006_evaluation} & 2004 & Local & Post-hoc & None & \multicolumn{2}{c|}{SF based on feature-utility} \\ \hline
        NUN \cite{cummins_2006_kleor,dasarathy_1995_nearest,doyle_2004_explanation} & 2006 & Local & Post-hoc & None & \multicolumn{2}{c|}{Natural CF} \\ \hline
        KLEOR \cite{cummins_2006_kleor} & 2006 & Local & Post-hoc & None & \multicolumn{2}{c|}{NUN-related SF} \\ \hline
        Nugent et al. \cite{nugent_2009_gaining} & 2009 & Local & Post-hoc & None & \multicolumn{2}{c|}{Local-region boundaries} \\ \hline
        \hline
        \textbf{INFLUENTIAL} & \multirow{2}{*}{Year} & Global / & Post-hoc / & \multicolumn{2}{c|}{Model or data} & Requires \\
        \textbf{INSTANCES} & & Local & Intrinsic & \multicolumn{2}{c|}{-type specificity} & model's gradients \\
        \hline
        Koh and Liang \cite{koh_2017} & 2017 & Both & Post-hoc & \multicolumn{2}{c|}{$\mathcal{L}$ twice-differentiable and strictly convex w.r.t. $\theta$} & Yes \\ \hline
        Khanna and al. \cite{khanna_2018} & 2018 & Local & Post-hoc & \multicolumn{2}{c|}{Requires an access to the function and gradient-oracles} & Yes \\ \hline
        Yeh and al. \cite{yeh_2018} & 2018 & Local & Intrinsic & \multicolumn{2}{c|}{Work for classification neural networks with regularization} & Yes \\ \hline
        Hara and al. \cite{hara_2019} & 2019 & Local & Intrinsic & \multicolumn{2}{c|}{Models trained with SGD, saving intermediate checkpoints} & Yes \\ \hline
        Koh and Liang \cite{koh_2019} & 2019 & Both & Post-hoc & \multicolumn{2}{c|}{$\mathcal{L}$ twice-differentiable and strictly convex w.r.t. $\theta$} & Yes \\ \hline
        Basu and al. \cite{basu_2019} & 2019 & Both & Post-hoc & \multicolumn{2}{c|}{$\mathcal{L}$ twice-differentiable and strictly convex w.r.t. $\theta$} & Yes \\ \hline
        Barshan and al. \cite{barshan_2020} & 2020 & Both & Post-hoc & \multicolumn{2}{c|}{$\mathcal{L}$ twice-differentiable and strictly convex w.r.t. $\theta$} & Yes \\ \hline
        Feldman and Zhang \cite{feldman_2020} & 2020 & Global & Intrinsic & \multicolumn{2}{c|}{Requires to train numerous models on subsampled datasets} & No \\ \hline
        Pruthi and al. \cite{pruthi_2020} & 2020 & Local & Intrinsic & \multicolumn{2}{c|}{Requires saving intermediate checkpoints} & Yes \\ \hline
        Sui and al. \cite{sui_2021} & 2021 & Local & Post-hoc & \multicolumn{2}{c|}{Work for classification neural networks} & Yes \\ \hline
        Chan and al. \cite{chen_2021} & 2021 & Both & Intrinsic & \multicolumn{2}{c|}{Requires saving intermediate checkpoints} & Yes \\ \hline
        \hline
        \multirow{2}{*}{\textbf{PROTOTYPES}} & \multirow{2}{*}{Year} & Global / & Post-hoc / & Model or data- & \multirow{2}{*}{Task} & \multirow{2}{*}{Other} \\
        & & Local & Intrinsic & type specificity & & \\
        \hline
        CCCD \cite{Priebe2003classification} & 2003 & Both & NA & by-design & Classification & Set cover \\ \hline
        ProtoSelect \cite{bien2011prototype} & 2011 & Both & NA & by-design & Classification & Set cover \\ \hline
        Kim et al. \cite{kim_bayesian_2015} & 2019 & Both & NA & by-design, tabular & Classification & Bayesian-based \\ \hline
        ProtoPNet \cite{chen2019looks} & 2019 & Both & NA & by-design, FGCV & Classification & Neural network \\ \hline
        ProSeNet \cite{ming2019interpretable} & 2019 & Both & NA & by-design, sequences & Classification & Neural network \\ \hline
        ProtoTree \cite{nauta2021neural} & 2021 & Both & NA & by-design, FGCV & Classification & Neural network \\ \hline
        M-PEER \cite{filho2023explainable} & 2023 & Both & Post-hoc & No & Regression & NA \\ \hline
        \hline 
        \multirow{2}{*}{\textbf{CONCEPTS}} & \multirow{2}{*}{Year} & Global / & Post-hoc / & Model or data & Need labeled & Concepts \\
        & & Local & Intrinsic & -type specificity & concepts & format\\
        \hline
        OPAM \cite{peng2017object} & 2017 & Global & NA & By-design, FGCV & Yes & part-based \\ \hline
        TCAV \cite{kim_2018_interpretability} & 2018 & Global & Post-hoc & Neural network & Yes & same as input \\ \hline
        ACE \cite{ghorbani_2019_towards} & 2019 & Global & Post-hoc & Neural network & No & segmented parts \\ \hline
        Zhang et al. \cite{zhang_2021_invertible} & 2021 & Global & Post-hoc & Neural network & No & segmented parts \\ \hline
        CRAFT \cite{fel_2022_craft} & 2022 & Global & Post-hoc & Neural network & No & crops \\ \hline
        Particul \cite{xu2022particul} & 2017 & Global & NA & By-design, FGCV & Yes & part-based \\ \hline
    \end{tabular}}
    \caption{Comparison table between the different natural example-based formats and methods. NA: Not applicable, FGCV: Fine-grained computer vision}
    \label{tab:methods_comparison}
    \end{table}

\section{Acknowledgments}
    This work has been supported by the French government under the “France 2030” program as part of the SystemX Technological Research Institute. This work was conducted as part of the Confiance.AI program, which aims to develop innovative solutions for enhancing the reliability and trustworthiness of AI-based systems. Additional funding was provided by ANR-3IA Artificial and Natural
    Intelligence Toulouse Institute (ANR-19-PI3A-0004).
    
    We are also thankful to the DEEL's\footnote{\href{https://www.deel.ai/}{https://www.deel.ai/}} core team for their expertise and feedback. A.M. Picard, D. Vigouroux, C. Friedrich, V. Mussot, and Y. Prudent.
    
    Finally, we are thankful to the authors who accepted our use of their figures. E.M Kenny and M.T. Keane \cite{kenny_2021_explaining,kenny_2021_generating}, F. Liu \cite{pruthi_2020}, B. Kim \cite{kim_examples_2016}, and T. Fel \cite{fel_2022_craft}.

\clearpage
\bibliographystyle{template/splncs04}

\end{document}